\documentclass[lettersize,journal]{IEEEtran}
\usepackage{amsmath,amsfonts}
\usepackage{algorithmic}
\usepackage{algorithm}
\usepackage{array}
\usepackage[caption=false,font=normalsize,labelfont=sf,textfont=sf]{subfig}
\usepackage{textcomp}
\usepackage{stfloats}
\usepackage{url}
\usepackage{verbatim}
\usepackage{booktabs}
\usepackage{graphicx}
\usepackage{cite}
\usepackage[table]{xcolor}
\usepackage{arydshln}
\usepackage{balance} 
\usepackage{multirow}
\begin{document}

\title{Finding the Needle in a Haystack: Test-Time Analog Circuit Representation Adaptation for Bayesian Optimization}

\author{Fin Amin, Sounak Dutta, and Paul D. Franzon%
\thanks{This paper was produced by the Department of Electrical and Computer Engineering, North Carolina State University, Raleigh, NC, USA.}%
\thanks{Manuscript received Month DD, 2026; revised Month DD, 2026.}}

\markboth{Journal of \LaTeX\ Class Files,~Vol.~14, No.~8, August~2021}%
{Amin \MakeLowercase{\textit{et al.}}: Finding a Needle in a Haystack: Test-Time Analog Circuit Representation Adaptation for Bayesian Optimization
}


\maketitle

\begin{abstract}
Bayesian optimization (BO) is a sample-efficient framework for analog circuit topology search, where evaluating each candidate topology can require costly simulation. However, representation-based BO methods typically treat circuit embeddings as fixed after encoder training. This creates a mismatch between representation learning and optimization: embeddings learned to encode or reconstruct circuit structure are not necessarily organized according to the figure of merit (FoM) being optimized. This paper introduces \underline{T}est-\underline{T}ime \underline{A}nalog \underline{R}epresentation Adaptation for Bayesian \underline{O}ptimization (TTARO), an online deep-kernel BO framework that adapts circuit representations throughout the search process. Starting from pretrained circuit embeddings, TTARO jointly learns a nonlinear feature transformation and a Gaussian-process surrogate using the FoM labels of the circuits evaluated so far. Following each new evaluation, TTARO updates the representation and surrogate before selecting the next candidate. We compare TTARO with conventional Gaussian Process-based BO over fixed embeddings and with Deep Kernel Learning (DKL), which learns the representation only from the initial evaluated designs and keeps it fixed throughout the remainder of the search. By continually incorporating newly observed FoM labels into representation learning, TTARO aligns the search space with the optimization objective as BO progresses. In our experiments, TTARO reduces regret AUC by 15.2\% on average relative to BO and by 20.7\% relative to DKL across 40 encoder/kernel/acquisition settings, outperforming prior art in most settings with reductions as large as 46.7\%.
\end{abstract}


\section{Introduction and Motivation}\label{sec:intro}
\IEEEPARstart{A}{nalog} circuit topology design remains a central challenge in electronic design automation because the search space is discrete, structured, and expensive to evaluate. A candidate topology must often be decoded, sized, simulated, and checked against design constraints before its quality is known. These costs make sample efficiency essential.

\begin{figure}[t]
    \centering
    \includegraphics[width=\columnwidth]{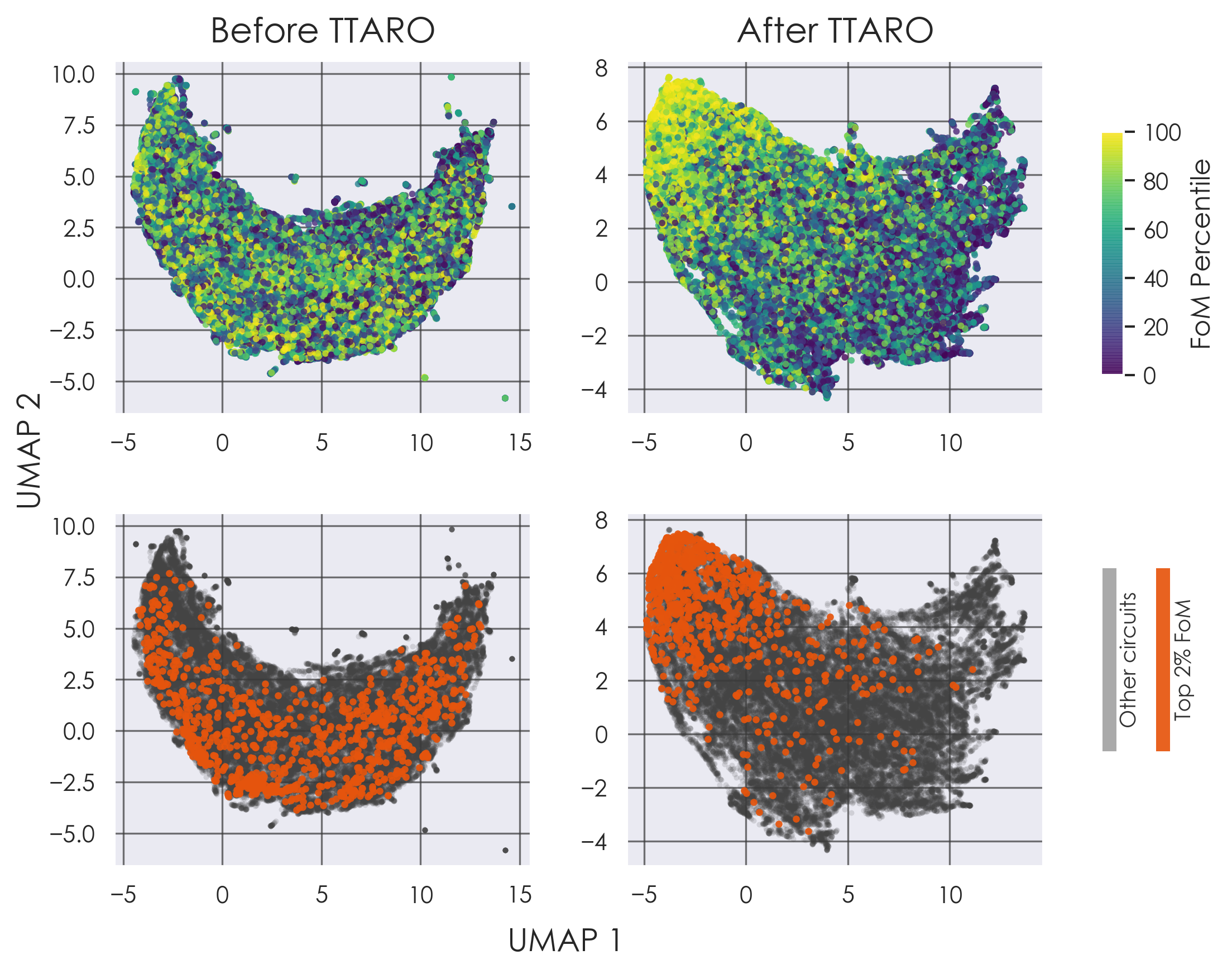}
    \caption{\textbf{Two-dimensional UMAP \cite{umap} visualization analog topology representations before and after TTARO adaptation, with each candidate colored by its FoM percentile.} Before adaptation, high- and low-performing circuits are broadly intermixed throughout the representation space. After TTARO, the FoM exhibits substantially stronger spatial organization, with high-performing candidates concentrated within a more coherent region. This objective-aligned geometry allows the Gaussian-process kernel to assign greater similarity to circuits with comparable performance, improving surrogate generalization and enabling the acquisition function to target promising regions more effectively.}
    \label{fig:ttaro_representation_geometry}
\end{figure}

Recent learning-assisted approaches encode circuit topologies into latent spaces and perform optimization over the resulting representations \cite{Shen2025ATOM,Dong2023CktGNN,Lu2022TopologyOptimizationGraphEmbedding}. This strategy makes topology search compatible with continuous black-box optimization methods such as Bayesian optimization (BO). Adjacent work further reflects the broader shift toward representation-centric analog design, including generative topology models such as AnalogGenie and circuit-level representation learning methods such as Ckt2Vec \cite{Gao2025AnalogGenie,Xu2025Ckt2Vec}. However, in latent-space BO frameworks, the learned representation is usually fixed before BO begins. As a result, the optimizer must operate in whatever geometry the encoder provides, even when that geometry is poorly aligned with the target FoM.

This paper introduces \underline{T}est-\underline{T}ime
\underline{A}nalog \underline{R}epresentation Adaptation for Bayesian \underline{O}ptimization (TTARO), an online deep-kernel Bayesian optimization framework that learns an objective-aware transformation of pretrained circuit embeddings as figure-of-merit (FoM) observations accumulate. At each BO iteration, TTARO jointly refits a nonlinear feature transformation and Gaussian-process (GP) surrogate using the circuits evaluated so far, then applies an arbitrary acquisition function to select the next candidate from the circuit bank. We compare TTARO against conventional GP-based BO over fixed embeddings, the celebrated Deep-Kernel Learning technique \cite{Wilson2016DeepKernelLearning}, which learns and freezes the transformation using only the initial evaluated designs, and an oracle fixed representation learned using all available FoM labels.

To investigate if TTARO improves optimization, we perform exhaustive experiments across permutations of multiple circuit encoders, kernels and acquisition functions, using two publicly available circuit benchmark search spaces.

The contributions of this paper are:
\begin{itemize}
\item We introduce Test-Time Analog Representation Adaptation, an online deep-kernel Bayesian optimization framework that repeatedly learns an objective-aware transformation of pretrained circuit embeddings as new FoM observations become available. As far as we are aware, no prior art has explored this framework for analog circuit topologies. 

\item We show that TTARO is a general BO framework compatible with expected improvement, upper confidence bound, and Thompson sampling, as well as linear and radial basis function Gaussian-process kernels.

\item We conduct a large-scale evaluation comprising 160 principal configurations across two public circuit-topology benchmarks, eight benchmark--encoder pairings, four representation regimes, and five kernel--acquisition settings, with each configuration evaluated over 20 random seeds.

\end{itemize}

\textbf{To the best of our knowledge, this is the most comprehensive evaluation ever performed concerning BO for analog circuit topology search.}

\section{Background and Related Work}

\subsection{Analog Circuit Topology Search}
Analog circuit design is commonly separated into topology selection and device sizing. Device sizing has received substantial attention because, once the topology is fixed, the remaining design variables can be treated as continuous parameters and optimized with simulation-driven methods. Bayesian optimization has been used in this setting to reduce the number of expensive circuit simulations, including neural-network-assisted BO for analog circuit synthesis \cite{Zhang2019BayesianOptimizationAnalogNN} and transfer-oriented BO for transistor sizing across circuit designs and technology nodes \cite{Xing2024KATO}. Related active-learning methods also treat simulation cost as the primary bottleneck and iteratively select circuit candidates whose evaluation is expected to improve the predictive model \cite{Guo2018ActiveLearningCircuitSynthesis,acof}.

Topology search is more difficult because candidate circuits are discrete, structured, and constrained by electrical validity. One line of work represents op-amp topologies at the behavioral level and searches over a lower-dimensional space. Lu et al. encode op-amp behavioral topologies as directed acyclic graphs, learn continuous embeddings with a variational graph autoencoder, and perform topology search in the embedding space using BO before decoding selected points back to circuit topologies \cite{Lu2022TopologyOptimizationGraphEmbedding}. ATOM extends this direction by defining a designer-comprehensible behavior-level op-amp design space, learning continuous topological representations, and using freeze-thaw BO to allocate simulation effort efficiently across candidate topologies \cite{Shen2025ATOM}. INTO-OA takes a different route: rather than forcing the topology space into a continuous latent space, it applies a Weisfeiler-Lehman (WL) graph kernel inside a GP surrogate and uses gradient information from the surrogate to identify performance-critical substructures and support interpretable topology refinement \cite{WL_Test, Shen2025INTOOA}. 

\begin{figure}[ht]
    \centering
    \includegraphics[
        width=\columnwidth,
        trim={0 0 0 30pt},
        clip
    ]{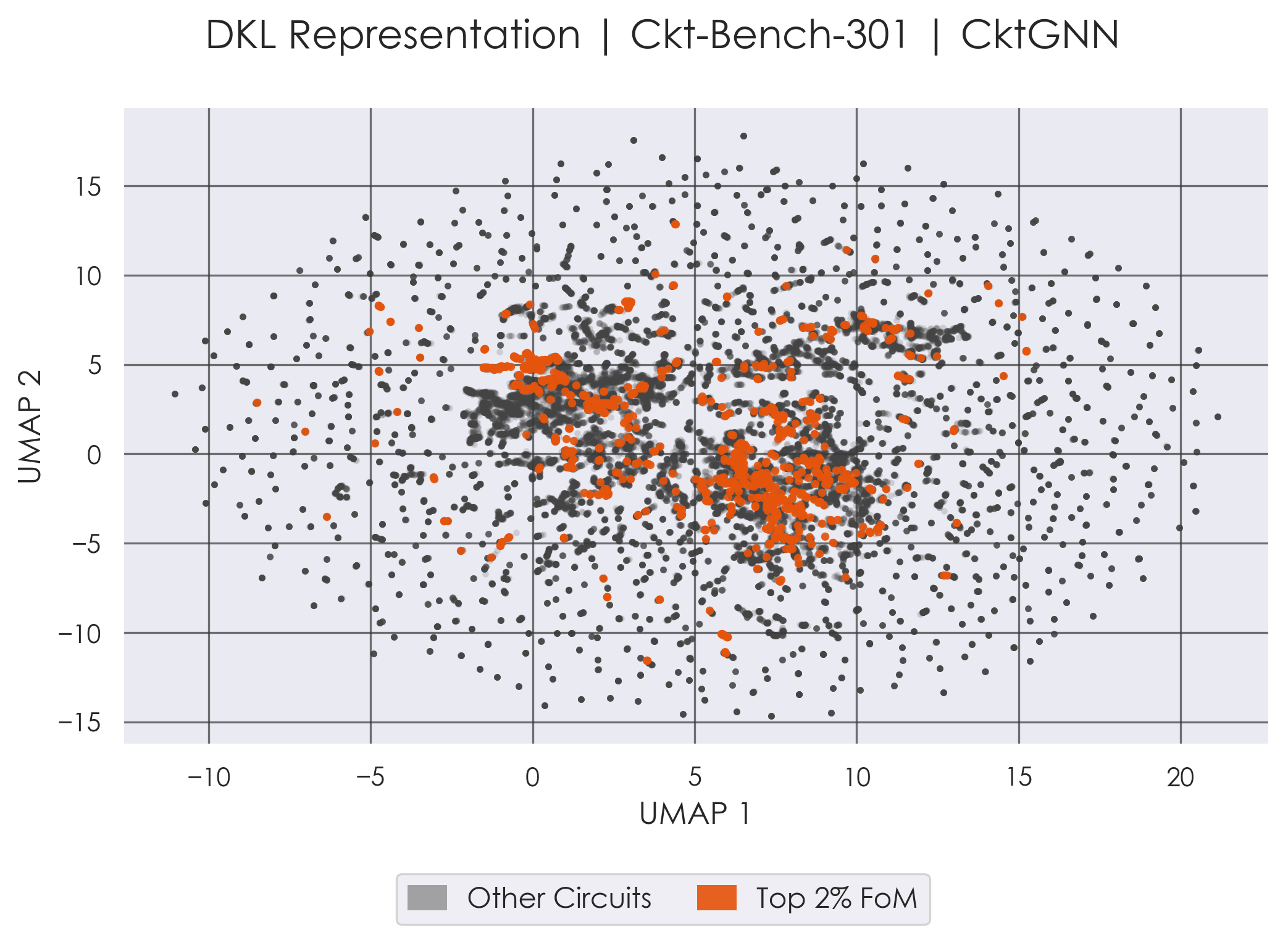}
    \caption{\textbf{Two-dimensional UMAP visualization of the DKL representation for Ckt-Bench-301 using CktGNN embeddings.} Circuits within the top 2\% of the FoM distribution are highlighted in orange, while all remaining candidates are shown in gray. Although DKL introduces localized performance structure, high- and low-performing circuits remain interspersed throughout much of the representation space. Because this representation is learned once and then frozen during DKL, it cannot incorporate the additional FoM evidence collected during Bayesian optimization. This motivates TTARO, which continually revises the surrogate geometry as new circuit evaluations become available.}
    \label{fig:dkl_top2_geometry}
\end{figure}

Recent work has also framed circuit topology design as graph or sequence generation. CktGNN represents circuits using a two-level graph neural network over a predefined subgraph basis and introduces the Open Circuit Benchmark for reproducible topology and sizing experiments over operational amplifiers \cite{Dong2023CktGNN}. AnalogGenie broadens the generative setting by building a larger analog-circuit topology dataset and representing circuits as pin-level graphs sequentialized through Eulerian circuits, enabling generation beyond a single op-amp topology family \cite{Gao2025AnalogGenie}. These methods show that learned circuit representations can support topology search, generation, and benchmarking, but they also create a new modeling question: how should an optimizer use a representation once circuit-level FoM observations begin to accumulate?

\subsection{Circuit Representation Learning}
Circuit representations must preserve both graph structure and circuit-relevant behavior. Generic graph encoders can capture connectivity, but analog circuits also contain directionality, functional substructures, device types, and continuous electrical characteristics. CktGNN addresses this by representing each circuit as a composition of subgraphs from a known basis and applying inner and outer GNNs to encode local circuit motifs and global directed message passing \cite{Dong2023CktGNN}. Related circuit-graph learning work has also studied representation quality for circuit equivalence tasks to detect circuit graph isomorphism efficiently \cite{dgn}. In op-amp topology optimization, graph autoencoders have been used to map discrete behavioral topologies to continuous representations suitable for surrogate modeling and search \cite{Lu2022TopologyOptimizationGraphEmbedding,Shen2025ATOM}. Graph-kernel methods provide an alternative representation route by measuring structural similarity directly in graph space, as in INTO-OA's WL-kernel surrogate \cite{Shen2025INTOOA}.

Other methods focus on making circuit embeddings more faithful to analog-domain semantics. Ckt2Vec argues that one-hot and text-based device encodings are limited because they do not reflect continuous electrical behavior; it instead extracts frequency-domain features from device I-V curves and combines these electrical features with graph contrastive learning for circuit-level representation learning \cite{Xu2025Ckt2Vec}. AnalogGenie similarly emphasizes representation fidelity, but from a generative perspective: by modeling device pins rather than whole devices as graph nodes, it avoids ambiguous mappings between generated graphs and valid circuit netlists \cite{Gao2025AnalogGenie}. Together, these works illustrate that the choice of representation is not a neutral preprocessing step. It determines which notions of circuit similarity are visible to downstream prediction, generation, and optimization models.

\begin{figure*}[t]
    \centering
    \includegraphics[width=\textwidth]
    {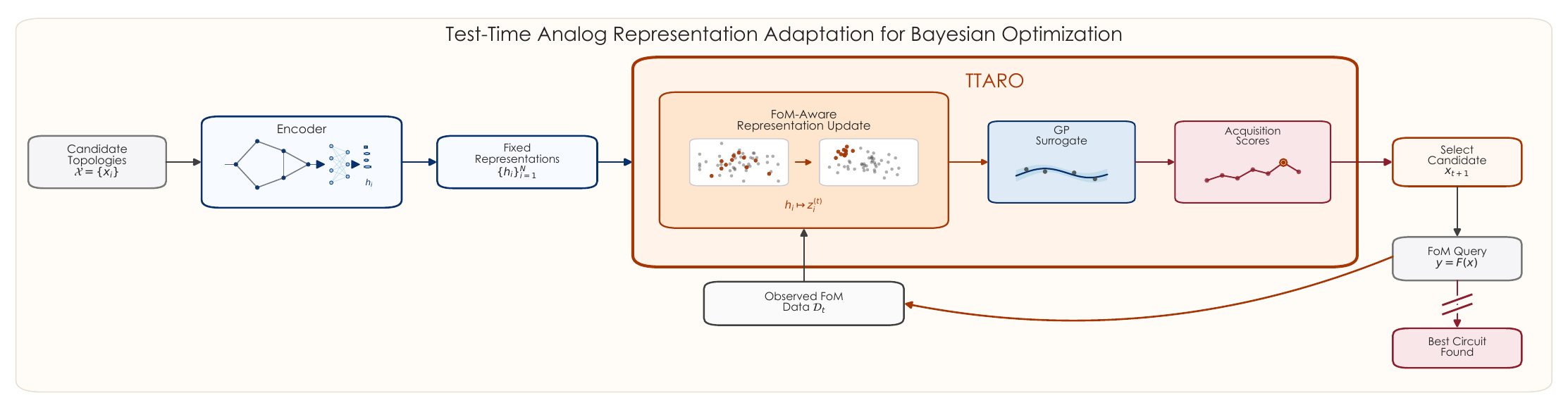}
    \caption{
        \textbf{Overview of Test-Time Analog Representation Adaptation for Bayesian
        Optimization (TTARO).} Candidate circuit topologies are first mapped
        by a pretrained encoder to fixed representations. At each BO
        iteration, TTARO uses the observed FoM data to adapt these
        representations, which are then used by the GP surrogate to compute
        acquisition scores and select the next circuit for evaluation. The
        newly observed FoM is added to the dataset, and the representation
        adaptation and BO procedure are repeated until the evaluation budget
        is exhausted. TTARO is agnostic to the choice of circuit encoder, acquisition function, and GP kernel. 
    }
    \label{fig:ttaro_overview}
\end{figure*}

\subsection{Bayesian Optimization in Learned Latent Spaces}
BO is attractive for analog design because it explicitly targets expensive black-box objectives with a small evaluation budget. A GP surrogate provides both a predictive mean and uncertainty estimate, while an acquisition function such as expected improvement (EI), upper confidence bound (UCB), or Thompson sampling (TS) chooses the next candidate to evaluate \cite{expectedImprovement,srinivas2009gaussian_UCB,thompsonSampling}. In analog design, this framework has been used for device sizing and synthesis tasks where each simulation is costly \cite{Zhang2019BayesianOptimizationAnalogNN,Xing2024KATO}. When the input is a structured object such as a circuit topology, BO is typically applied after mapping the object into a continuous feature or latent space \cite{Lu2022TopologyOptimizationGraphEmbedding,Dong2023CktGNN}.

The effectiveness of this procedure depends strongly on the surrogate's coordinate system and kernel. Input warping shows that learning transformations of the input space can make GP-based BO more effective on nonstationary objectives \cite{Snoek2014InputWarpingBO}. Deep kernel learning (DKL) generalizes this idea by composing a neural feature map with a GP kernel, allowing the model to learn a representation and kernel jointly through the GP marginal likelihood \cite{Wilson2016DeepKernelLearning}. These methods motivate learned transformations that both represent structured inputs and improve the surrogate used for sequential decision-making.

\subsection{Adaptive Representations for Bayesian Optimization} \label{sec:adaptive_reps4BO}
Several BO methods explicitly study how representation learning affects optimization. Deep Kernel Bayesian Optimization applies DKL directly inside BO, learning a deep kernel during optimization so that the GP operates in a feature space better suited to the observed objective \cite{Bowden2021DeepKernelBayesianOptimization}. SILBO learns a low-dimensional embedding iteratively using both labeled evaluations and unlabeled candidate points proposed by the acquisition function, addressing the difficulty of high-dimensional BO when a single initial projection is inadequate \cite{Chen2020SILBO}. For structured search spaces, contrastive embedding methods use known structural relationships, such as subtree replacements in grammar-defined objects, to learn continuous spaces in which BO can exploit local similarity more effectively \cite{Tingey2021ContrastiveEmbeddingStructuredBO}. LOCo identifies collisions in learned latent spaces, where points with very different objective values are mapped too close together, and introduces a regularizer to discourage such collisions during BO \cite{Zhang2022LOCo}. CoBO similarly targets latent-space quality by encouraging correlation between latent-space distances and objective-value differences, with additional weighting around promising regions of the search space \cite{Lee2023CoBO}.

Taken together, these methods show that representation quality during BO has been studied across a range of structured and scientific optimization settings. Deep Kernel Bayesian Optimization evaluates adaptive deep kernels on protein engineering, antibody design, and nanophotonics tasks \cite{Bowden2021DeepKernelBayesianOptimization}. SILBO studies iterative embedding learning for high-dimensional BO and hyperparameter optimization \cite{Chen2020SILBO}. Contrastive embedding and CoBO show that latent-space geometry matters in structured search spaces such as molecule design and symbolic expression optimization \cite{Tingey2021ContrastiveEmbeddingStructuredBO,Lee2023CoBO}. LOCo further shows that collisions in learned latent spaces can degrade BO performance and introduces a regularizer for dynamic embedding-based BO \cite{Zhang2022LOCo}. LaMBO couples a denoising autoencoder with a multi-task GP head for latent-space BO in small-molecule and fluorescent-protein design \cite{stanton2022accelerating}, and DKL has also been used as a BO surrogate for chemical reaction outcome optimization \cite{reactionOutComeDKL}. TTARO brings this representation-adaptive view to finite-bank analog topology search, where pretrained circuit embeddings are adapted as FoM labels are observed.

\section{Problem Formulation and Setting}\label{sec:problem_setting}

Recent analog topology-search methods often begin by constructing or enumerating a set of candidate circuit topologies, then mapping those candidates into a representation space suitable for learning or optimization. For example, topology-generation and topology-optimization frameworks define behavior-level or graph-based op-amp design spaces, encode each topology as a DAG or circuit graph, and use learned continuous representations to make the discrete topology space searchable by BO \cite{Lu2022TopologyOptimizationGraphEmbedding,Shen2025ATOM}. Benchmark-driven methods such as OCB similarly provide finite banks of valid candidate circuits together with graph-based circuit representations and precomputed performance labels \cite{Dong2023CktGNN}. In this paper, we assume this upstream topology-generation or benchmark-construction step has already produced a finite candidate bank.

Let
\begin{equation}
    \mathcal{X} = \{x_1, x_2, \ldots, x_N\}
\end{equation}
denote this library of candidate analog circuit topologies. Each $x_i$ may be viewed as a circuit graph, netlist, or topology-level design instance. Each candidate has an associated FoM
\begin{equation}
    y_i = F(x_i),
\end{equation}
which is unknown to the optimizer until that circuit is evaluated. In a deployed circuit-design loop, this evaluation would correspond to simulation, sizing, or measurement. In our benchmark setting, the evaluation is implemented as a label query: selecting a circuit reveals its stored FoM. Given an evaluation budget $B \ll N$, the objective is to identify
\begin{equation}
    x^\star
    =
    \underset{x_i \in \mathcal{X}}{\operatorname{arg\,max}}
    \; F(x_i)
\end{equation}
using as few circuit evaluations as possible.

Each candidate is also associated with a fixed initial representation
\begin{equation}
    h_i \in \mathbb{R}^{d_h},
\end{equation}
computed before BO begins. This representation may come from a circuit-specific encoder, a graph autoencoder, or a structural graph feature map. Its purpose is to provide coordinates in which a surrogate model can compare circuits: circuits that are close in representation space are treated as similar by the kernel, and observations from evaluated circuits are generalized to unevaluated circuits through that geometry. However, these representations are typically trained to encode circuit structure, reconstruct graphs, or capture topology semantics, not to organize circuits according to the specific FoM being optimized in the current run.

At BO iteration $t$, let
\begin{equation}
    \mathcal{I}_t \subseteq \{1,2,\ldots,N\}
\end{equation}
denote the indices of the circuits evaluated so far. The observed dataset is
\begin{equation}
    \mathcal{D}_t
    =
    \left\{
        (h_i,y_i)
        \mid
        i \in \mathcal{I}_t
    \right\}.
\end{equation}
All remaining candidates are unevaluated but already have representations. The optimizer must therefore use the observed FoMs and the representation geometry to decide which unevaluated circuit should be queried next.

Conventional representation-based BO fits a GP surrogate directly over the fixed representations:
\begin{equation}
    f(h)
    \sim
    \mathcal{GP}
    \left(
        m(h),
        k(h,h')
    \right).
\end{equation}
In this setting, the kernel defines circuit similarity entirely through the original representation space. If the encoder places two circuits nearby because they are structurally similar but their FoMs differ substantially, the surrogate can make misleading predictions; conversely, high-performing circuits that are far apart in the initial space may not share useful statistical strength. TTARO addresses this mismatch by learning a FoM-informed transformation of the fixed input representations throughout the BO process, so that the surrogate geometry becomes increasingly aligned with the objective as circuit evaluations accumulate.

\section{Test-Time Analog Representation Adaptation}\label{sec:ttaro_alg}

TTARO adapts a fixed circuit representation during BO by learning a feature map from the FoM observations collected so far. Algorithm~\ref{alg:ttaro} summarizes the full loop: TTARO normalizes the candidate representation bank, initializes an evaluated set, repeatedly fits a deep-kernel surrogate, scores unevaluated candidates, and adds the newly evaluated circuit to the observed dataset. At iteration $t$, each initial circuit representation is first normalized feature-wise over the candidate bank:
\begin{equation}
    \bar{h}_{i,q}
    =
    \frac{
        h_{i,q} - h_q^{\min}
    }{
        h_q^{\max} - h_q^{\min}
    },
\end{equation}
where $h_q^{\min}$ and $h_q^{\max}$ are computed over all candidates in $\mathcal{X}$. TTARO then maps the normalized representation into a task-adapted latent space:
\begin{equation}
    z_i^{(t)}
    =
    \phi_{\theta_t}(\bar{h}_i),
    \qquad
    z_i^{(t)} \in \mathbb{R}^{d_z}.
    \label{eq:ttaro_feature_map}
\end{equation}
Here, $\theta_t$ denotes the feature-map parameters learned from the FoM observations available at iteration $t$.

\begin{algorithm}[hb]
\caption{Test-Time Analog Circuit \\
Representation Adaptation (TTARO)}
\label{alg:ttaro}
\begin{algorithmic}[1]
\STATE \textbf{Input:} Candidate representations $\{h_i\}_{i=1}^{N}$ \label{alg:ttaro:input_repr}
\STATE \textbf{Input:} Initial sample size $n_0$ and evaluation budget $B$ \label{alg:ttaro:input_budget}
\STATE \textbf{Input:} Acquisition function $a$ \label{alg:ttaro:input_acq}
\STATE Normalize the candidate representation bank \label{alg:ttaro:normalize}
\STATE Randomly select an initial index set $\mathcal{I}_{n_0}$ with $|\mathcal{I}_{n_0}|=n_0$ \label{alg:ttaro:init_set}
\STATE Evaluate the circuits in $\mathcal{I}_{n_0}$ \label{alg:ttaro:init_eval}
\STATE Construct the initial dataset $\mathcal{D}_{n_0}$ \label{alg:ttaro:init_data}
\FOR{$t=n_0$ to $B-1$}
    \STATE Initialize the feature map and exact GP \label{alg:ttaro:init_model}
    \STATE Train the feature map and GP using $\mathcal{D}_t$ \label{alg:ttaro:train}
    \STATE Compute acquisition scores for all $i \notin \mathcal{I}_t$ \label{alg:ttaro:score}
    \STATE Select $i_{t+1}=\arg\max_{i \notin \mathcal{I}_t} a_t(h_i)$ \label{alg:ttaro:select}
    \STATE Evaluate circuit $i_{t+1}$ and observe $y_{i_{t+1}}$ \label{alg:ttaro:evaluate}
    \STATE $\mathcal{I}_{t+1}=\mathcal{I}_t \cup \{i_{t+1}\}$
    \STATE $\mathcal{D}_{t+1}=\mathcal{D}_t \cup \{(\bar{h}_{i_{t+1}},y_{i_{t+1}})\}$ \label{alg:ttaro:update}
\ENDFOR
\STATE \textbf{return} The best evaluated circuit in $\mathcal{I}_B$ \label{alg:ttaro:return}
\end{algorithmic}
\end{algorithm}

The feature map used in our experiments is a two-layer multilayer perceptron:
\begin{equation}
    \phi_{\theta_t}(\bar{h})
    =
    W_2
    \operatorname{Dropout}
    \left(
        \operatorname{ReLU}
        \left(
            W_1 \bar{h} + b_1
        \right)
    \right)
    +
    b_2.\label{eq:mlp}
\end{equation}
The hidden layer has width 128, and the transformed representation has dimension $d_z=16$.

A GP surrogate is defined over the transformed representations:
\begin{equation}
    f(h)
    \sim
    \mathcal{GP}
    \left(
        m_{\eta_t}
        \left(
            \phi_{\theta_t}(\bar{h})
        \right),
        k_{\psi_t}
        \left(
            \phi_{\theta_t}(\bar{h}),
            \phi_{\theta_t}(\bar{h}')
        \right)
    \right).
    \label{eq:ttaro_deep_kernel_gp}
\end{equation}
The experiments use either a scaled linear kernel,
\begin{equation}
    k_{\psi_t}^{\mathrm{linear}}(z,z')
    =
    \sigma_{f,t}^{2} z^\top z',
    \label{eq:ttaro_linear_kernel}
\end{equation}
or a scaled radial basis function (RBF) kernel,
\begin{equation}
    k_{\psi_t}^{\mathrm{RBF}}(z,z')
    =
    \sigma_{f,t}^{2}
    \exp
    \left(
        -\frac{1}{2}
        \sum_{q=1}^{d_z}
        \frac{
            (z_q-z'_q)^2
        }{
            \ell_{t,q}^{2}
        }
    \right).
    \label{eq:ttaro_rbf_kernel}
\end{equation}
The parameters $\psi_t$ include the kernel output scale, noise term, and, for the RBF kernel, the length-scale parameters.

\begin{figure*}[ht]
    \centering
    \includegraphics[
        width=\textwidth,
        trim={0 0pt 0 30pt},
        clip
    ]{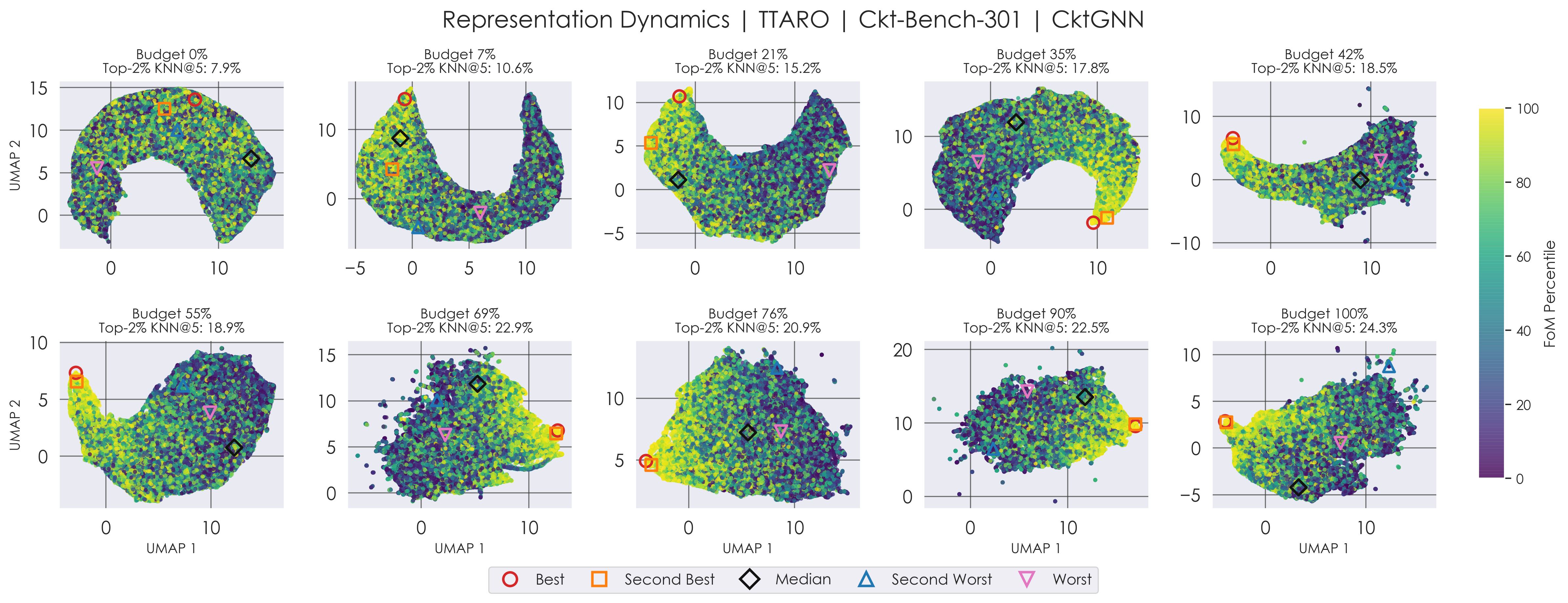}
    \caption{\textbf{Two-dimensional UMAP visualization of analog topology representations during TTARO adaptation on Ckt-Bench-301 using CktGNN embeddings.} Each candidate is colored by its FoM percentile. Panel subtitles report Top-2\% KNN@5, the fraction of five nearest representation-space neighbors of top-2\% FoM circuits that are also top-2\% circuits. As TTARO progresses, high-FoM candidates become more locally concentrated, indicating stronger objective-aligned representation geometry.}
    \label{fig:ttaro_representation_dynamics}
\end{figure*}

\subsection{Joint Representation and Surrogate Learning}

Lines~\ref{alg:ttaro:init_model}--\ref{alg:ttaro:train} of Algorithm~\ref{alg:ttaro} correspond to the surrogate-fitting step. At BO iteration $t$, TTARO trains the feature-map parameters $\theta_t$ together with the GP hyperparameters $\psi_t$ and mean parameters $\eta_t$ using the evaluated dataset $\mathcal{D}_t$. Let $\mathcal{I}_t = \{i_1,\ldots,i_{n_t}\}$ denote the evaluated circuit indices. The transformed training representations are collected as
\begin{equation}
    \mathbf{Z}_t
    =
    \left[
        z_{i_1}^{(t)},
        z_{i_2}^{(t)},
        \ldots,
        z_{i_{n_t}}^{(t)}
    \right]^\top .
\end{equation}
The observed FoMs are standardized to form $\tilde{\mathbf{y}}_t$. The feature map and GP are fit by minimizing the negative log marginal likelihood:
\begin{equation}
    \mathcal{L}_t
    =
    \frac{1}{2}
    \tilde{\mathbf{y}}_t^\top
    \mathbf{K}_t^{-1}
    \tilde{\mathbf{y}}_t
    +
    \frac{1}{2}
    \log
    \left|
        \mathbf{K}_t
    \right|
    +
    \frac{n_t}{2}
    \log 2\pi .
    \label{eq:ttaro_nll}
\end{equation}
The covariance matrix $\mathbf{K}_t$ is computed from the transformed representations in $\mathbf{Z}_t$. Its entries are
\begin{equation}
    \mathbf{K}_t[a,b]
    =
    k_{\psi_t}
    \left(
        z_{i_a}^{(t)},
        z_{i_b}^{(t)}
    \right)
    +
    \sigma_{n,t}^{2}\mathbf{1}[a=b].
    \label{eq:ttaro_training_covariance}
\end{equation}
Training uses only FoM observations from circuits already selected by BO. After fitting, TTARO applies the learned feature map to the full candidate bank, computes the GP posterior over unevaluated circuits, and evaluates the acquisition function to select the next circuit.

\subsection{Online Representation Adaptation}

\begin{figure*}[ht]
    \centering
    \includegraphics[
        width=\textwidth,
        trim={0 0 0 40pt},
        clip
    ]{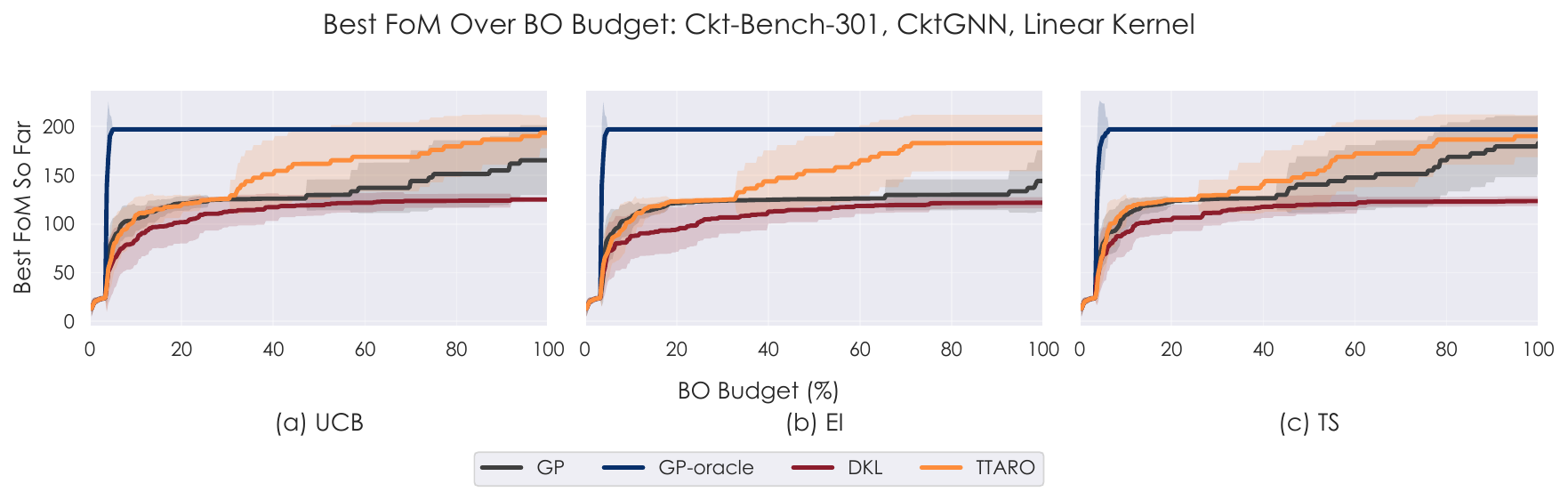}
    \caption{
        \textbf{Best observed FoM as a function of BO budget on
        Ckt-Bench-301 using CktGNN representations and a linear kernel.}
        The acquisition functions are (a) upper confidence bound (UCB), (b) expected improvement
        (EI), and (c) Thompson sampling (TS).
        The curves compare GP, GP-oracle, DKL, and TTARO, while the shaded
        regions indicate one standard deviation across runs. 
    }
    \label{fig:ckt301_cktgnn_linear_acquisitions}
\end{figure*}

Lines~\ref{alg:ttaro:score}--\ref{alg:ttaro:update} of Algorithm~\ref{alg:ttaro} describe the online BO step. After fitting the deep-kernel surrogate, an acquisition function $a_t$ is evaluated for every unevaluated candidate. The next circuit is selected according to
\begin{equation}
    i_{t+1}
    =
    \underset{
        i \notin \mathcal{I}_t
    }{
        \operatorname{arg\,max}
    }
    \;
    a_t(h_i).
\end{equation}
The selected circuit is evaluated, and its FoM is added to the observed set:
\begin{equation}
    \mathcal{I}_{t+1}
    =
    \mathcal{I}_t
    \cup
    \{i_{t+1}\}.
\end{equation}
After the update in Line~\ref{alg:ttaro:update}, the loop returns to Line~\ref{alg:ttaro:init_model}: TTARO retrains the feature map and GP using all FoM observations accumulated so far. The procedure is agnostic to the acquisition function used to score unevaluated candidates.

\subsection{Distinction from Deep Kernel Learning}

The DKL baseline uses the same deep-kernel surrogate structure, with the feature map trained only on the initial evaluated dataset. This produces an initial transformed representation bank
\begin{equation}
    \mathcal{Z}^{(0)}
    =
    \left\{
        \phi_{\theta_0}(\bar{h}_i)
    \right\}_{i=1}^{N}.
\end{equation}
This transformed representation bank remains fixed for the rest of the optimization process. Additional FoM observations update the GP posterior, while the representation supplied to the GP stays unchanged.

TTARO learns a sequence of feature maps
\begin{equation}
    \phi_{\theta_0},
    \phi_{\theta_1},
    \ldots,
    \phi_{\theta_{B-1}},
\end{equation}
using the growing set of evaluated circuits. The comparison between DKL and TTARO therefore isolates the effect of updating the representation used by the surrogate during the BO process.

\subsection{On the Importance of Aligned Kernels}

The central assumption behind GP-based BO is that the kernel encodes a useful notion of similarity for the objective being modeled. In this work, the relevant objective is the circuit FoM. The kernel should therefore assign high similarity to circuits with similar FoM values and lower similarity to circuits with substantially different FoM values.

Equations~\eqref{eq:ttaro_feature_map}--\eqref{eq:ttaro_rbf_kernel} show how TTARO changes the covariance structure used by the GP. Substituting the learned representation $z_i^{(t)}=\phi_{\theta_t}(\bar{h}_i)$ into the linear kernel gives
\begin{equation}
    k_{\psi_t}^{\mathrm{linear}}(x_i,x_j)
    =
    \sigma_{f,t}^{2}
    \phi_{\theta_t}(\bar{h}_i)^\top
    \phi_{\theta_t}(\bar{h}_j).
    \label{eq:ttaro_linear_substitution}
\end{equation}
Thus, the linear-kernel similarity between two circuits is the scaled inner product of their TTARO-adapted representations. Updating $\theta_t$ rotates, rescales, and reshapes the feature coordinates that determine this inner product.

For the RBF kernel, the same substitution gives
\begin{equation}
    k_{\psi_t}^{\mathrm{RBF}}(x_i,x_j)
    =
    \sigma_{f,t}^{2}
    \exp
    \left(
        -\frac{1}{2}
        \sum_{q=1}^{d_z}
        \frac{
            \left(
                \phi_{\theta_t,q}(\bar{h}_i)
                -
                \phi_{\theta_t,q}(\bar{h}_j)
            \right)^2
        }{
            \ell_{t,q}^{2}
        }
    \right),
    \label{eq:ttaro_rbf_substitution}
\end{equation}

where $\phi_{\theta_t,q}(\cdot)$ denotes the $q$th coordinate of the learned representation. In this case, updating $\theta_t$ changes the pairwise distances that appear inside the exponential. Circuits placed closer together in the adapted space receive larger covariance, while circuits separated along dimensions with small length scales receive smaller covariance.

These pairwise similarities enter the GP through the training covariance matrix. For the linear kernel, Eq.~\eqref{eq:ttaro_training_covariance} becomes
\begin{equation}
    \mathbf{K}_t^{\mathrm{linear}}[a,b]
    =
    \sigma_{f,t}^{2}
    \phi_{\theta_t}(\bar{h}_{i_a})^\top
    \phi_{\theta_t}(\bar{h}_{i_b})
    +
    \sigma_{n,t}^{2}\mathbf{1}[a=b].
    \label{eq:ttaro_linear_K}
\end{equation}
For the RBF kernel, the corresponding entry is
\begin{equation}
\begin{aligned}
    \mathbf{K}_t^{\mathrm{RBF}}[a,b]
    &=
    \sigma_{f,t}^{2}
    \exp
    \Bigg(
        -\frac{1}{2}
        \sum_{q=1}^{d_z}
        \frac{
            \Delta_{abq,t}^{2}
        }{
            \ell_{t,q}^{2}
        }
    \Bigg)
    \\
    &\quad
    +
    \sigma_{n,t}^{2}\mathbf{1}[a=b],
\end{aligned}
\label{eq:ttaro_rbf_K}
\end{equation}
where
\begin{equation}
    \Delta_{abq,t}
    =
    \phi_{\theta_t,q}(\bar{h}_{i_a})
    -
    \phi_{\theta_t,q}(\bar{h}_{i_b}) .
\end{equation}
Equations~\eqref{eq:ttaro_linear_K} and~\eqref{eq:ttaro_rbf_K} make the dependence on $\theta_t$ explicit. The learned feature map determines every off-diagonal entry of $\mathbf{K}_t$, which means it controls how FoM observations from one evaluated circuit influence the inferred FoM of other evaluated circuits.

The same dependence appears in predictions for unevaluated candidates. For an unevaluated circuit $x_j$, define the cross-covariance vector
\begin{equation}
    \mathbf{k}_{t,j}
    =
    \left[
        k_{\psi_t}(x_j,x_{i_1}),
        \ldots,
        k_{\psi_t}(x_j,x_{i_{n_t}})
    \right]^\top .
    \label{eq:ttaro_cross_covariance}
\end{equation}
For the linear kernel, its $a$th entry is
\begin{equation}
    \mathbf{k}_{t,j}^{\mathrm{linear}}[a]
    =
    \sigma_{f,t}^{2}
    \phi_{\theta_t}(\bar{h}_{j})^\top
    \phi_{\theta_t}(\bar{h}_{i_a}).
    \label{eq:ttaro_linear_cross_covariance}
\end{equation}
For the RBF kernel, its $a$th entry is
\begin{equation}
    \mathbf{k}_{t,j}^{\mathrm{RBF}}[a]
    =
    \sigma_{f,t}^{2}
    \exp
    \left(
        -\frac{1}{2}
        \sum_{q=1}^{d_z}
        \frac{
            \left(
                \phi_{\theta_t,q}(\bar{h}_{j})
                -
                \phi_{\theta_t,q}(\bar{h}_{i_a})
            \right)^2
        }{
            \ell_{t,q}^{2}
        }
    \right).
    \label{eq:ttaro_rbf_cross_covariance}
\end{equation}
The posterior mean and variance used by the acquisition function are
\begin{align}
    \mu_t(x_j)
    &=
    m_t(x_j)
    +
    \mathbf{k}_{t,j}^{\top}
    \mathbf{K}_t^{-1}
    \left(
        \mathbf{y}_t - \mathbf{m}_t
    \right), \\
    \sigma_t^2(x_j)
    &=
    k_{\psi_t}(x_j,x_j)
    -
    \mathbf{k}_{t,j}^{\top}
    \mathbf{K}_t^{-1}
    \mathbf{k}_{t,j}.
\end{align}

The prediction for $x_j$ is therefore controlled by two learned similarity objects: $\mathbf{K}_t$, which describes similarity among evaluated circuits, and $\mathbf{k}_{t,j}$, which describes similarity between the unevaluated candidate and the evaluated set. Updating $\theta_t$ changes both objects. This directly changes the posterior mean, the posterior variance, and the acquisition values used to select the next circuit. TTARO uses the accumulating FoM labels to continually revise this kernel geometry, making the surrogate increasingly aligned with the objective being optimized.

\begin{figure*}[ht]
    \centering
    \subfloat[\footnotesize Ckt-Bench-101.\label{fig:auc_by_scenario_101}]{%
        \includegraphics[
            width=0.49\textwidth,
            trim={0 0 0 25pt},
            clip
        ]{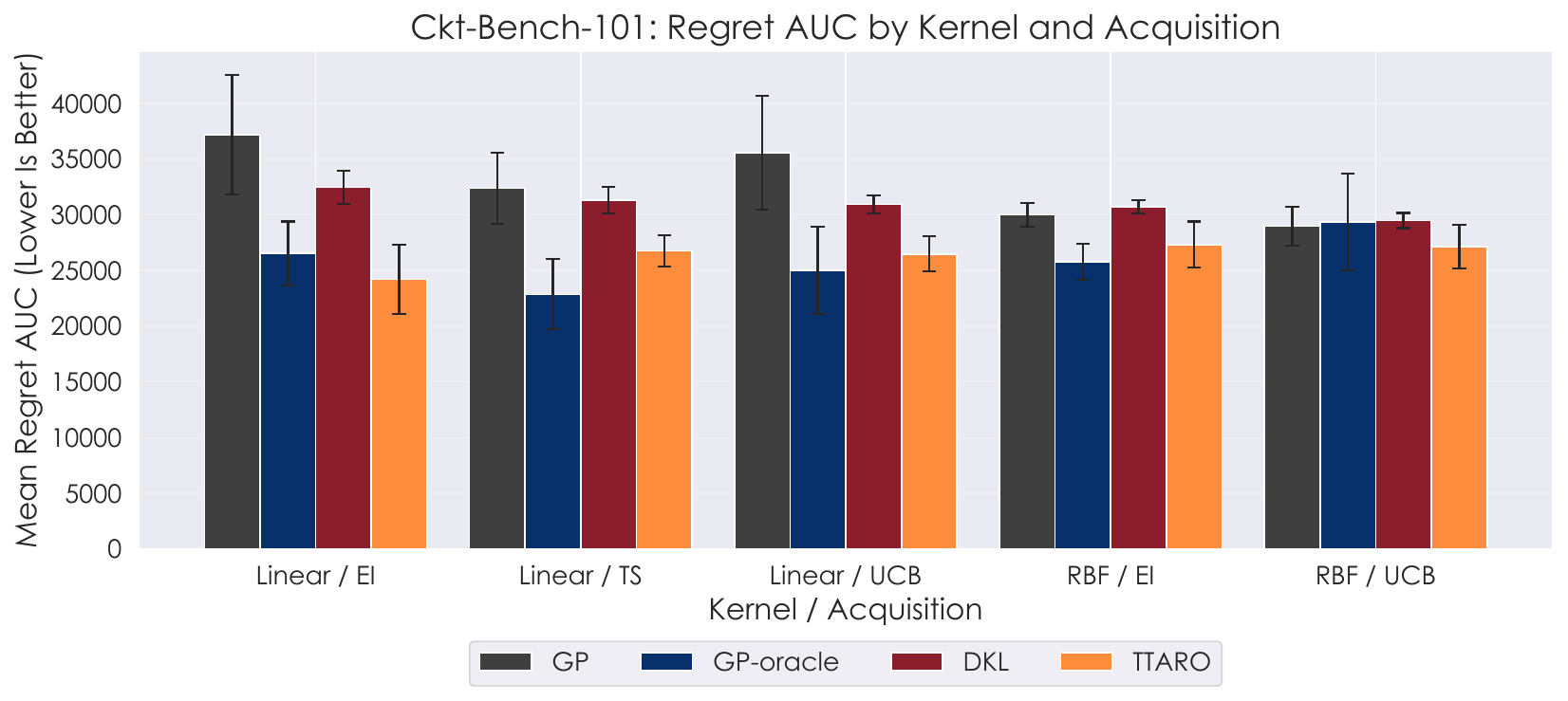}%
    }\hfill%
    \subfloat[\footnotesize Ckt-Bench-301.\label{fig:auc_by_scenario_301}]{%
        \includegraphics[
            width=0.49\textwidth,
            trim={0 0 0 25pt},
            clip
        ]{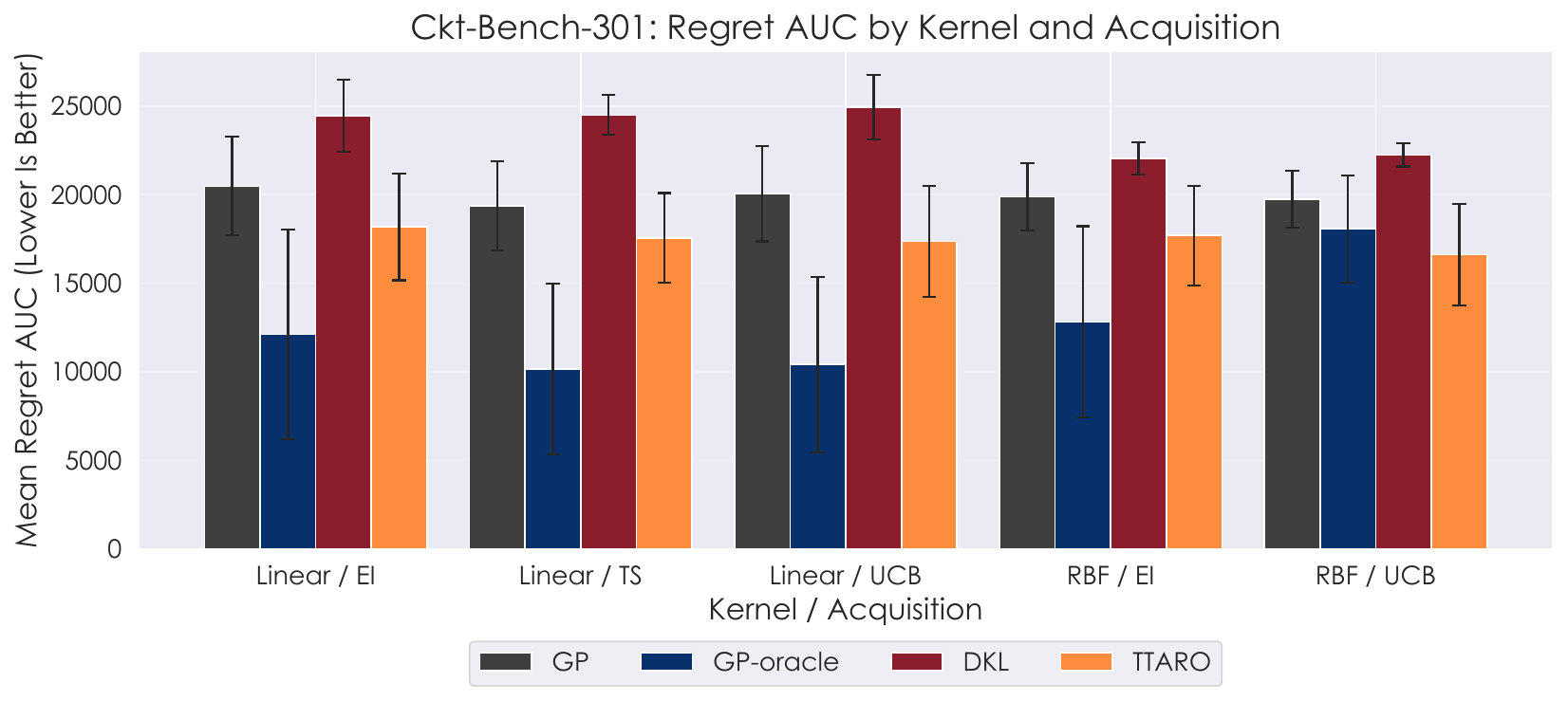}%
    }

    \caption{\textbf{Mean Regret AUC Across Kernel and Acquisition-Function Configurations.} Lower values indicate more sample-efficient optimization. Error bars denote the standard error of the mean across encoders.}
    \label{fig:auc_encoder}
\end{figure*}

\section{Experimental Setup}\label{sec:experiments}

\subsection{Benchmark Search Spaces}
We evaluate TTARO on the two operational-amplifier topology libraries from the Open Circuit Benchmark (OCB) \cite{Dong2023CktGNN}. Ckt-Bench-101 contains 10,000 valid circuit candidates, while Ckt-Bench-301 contains 50,000 candidates. Each candidate is associated with circuit measurements, including gain, phase margin, bandwidth, validity, and FoM. 

We use the default OCB FoM definition so that the reported results remain directly comparable with future work on the same benchmark. The FoM is computed as a weighted combination of normalized gain, phase margin, and bandwidth:
\begin{equation}
    \mathrm{FoM}
    =
    1.2\frac{|\mathrm{Gain}|}{100}
    +
    1.6\frac{\mathrm{PM}}{-90}
    +
    10\frac{|\mathrm{BW}|}{10^9}.
\end{equation}
During BO, selecting a candidate reveals its stored FoM. This finite-bank label-query setting allows all methods to be compared using the same candidate set, objective values, and evaluation budget.

\subsection{Circuit Representations}
For each benchmark candidate $x_i$, TTARO takes a fixed vector representation $h_i$ as input. The reported representation sets are dataset-specific. For Ckt-Bench-101, we evaluate CktGNN, D-VAE, D-VAE-GCN, and WL representations. For Ckt-Bench-301, we evaluate CktGNN, DAGNN, D-VAE, and D-VAE-GCN representations.

CktGNN is the circuit-specific encoder proposed with OCB; it represents a circuit using a predefined subgraph basis and applies a two-level GNN to encode local substructures and global circuit connectivity \cite{Dong2023CktGNN}. DAGNN provides a directed-acyclic-graph neural encoder baseline \cite{dagnn}. D-VAE and D-VAE-GCN provide graph autoencoding baselines, with D-VAE-GCN replacing the base message-passing component with graph-convolutional layers \cite{dvae,gcn}. The WL representation is a fixed structural graph feature map rather than a learned neural embedding \cite{WL_Test}.

The learned encoders produce 66-dimensional vectors, giving $10000 \times 66$ representation banks on Ckt-Bench-101 and $50000 \times 66$ representation banks on Ckt-Bench-301 for the corresponding encoder/dataset pairs. The WL representation used for Ckt-Bench-101 has 1,838 dimensions. This representation set tests TTARO across compact learned circuit embeddings and a substantially higher-dimensional structural feature map.

\subsection{Compared Methods}
We compare methods that differ in how the representation used by the GP surrogate is treated during BO. All methods start from the same pretrained circuit embeddings and operate over the same finite candidate bank.

The first baseline is a fixed-representation Gaussian process (\textbf{GP})--the technique used by the authors of OCB. In this setting, the original embedding $h_i$ is used directly as the GP input for every candidate $x_i$. As new FoM labels are observed, only the GP posterior is updated; the representation itself remains unchanged. 

The second baseline is \textbf{DKL} \cite{Wilson2016DeepKernelLearning}. The DKL feature map is trained using only the initial evaluated set, producing transformed representations $\phi_{\theta_0}(h_i)$. For parity, we implement DKL's feature map identically to TTARO via equation \ref{eq:mlp}. This transformation is then frozen and applied to the full candidate bank for the remainder of the BO run. Subsequent evaluations update the GP posterior, but do not update the feature map.

\textbf{TTARO} uses the same deep-kernel surrogate structure, but keeps representation learning coupled to the online search process. After each new circuit evaluation, TTARO retrains the feature map and GP using all FoM observations collected so far. Thus, the representation used by the surrogate can adapt as the evaluated set grows.

The oracle representation (\textbf{GP-oracle}) is included as an upper-reference condition. It trains the feature map using all FoM labels in the candidate bank before BO begins, including labels that would not be available in a real search. This condition is not a deployable optimization method; it measures how much improvement could be expected if the representation were fully aligned with the objective.

\begin{table*}[ht]
\centering
\caption{Full configuration-level Bayesian optimization performance on Ckt-Bench-101. Each row corresponds to one encoder, kernel, and acquisition-function setting. Regret AUC is reported as mean $\pm$ standard error; lower is better. The best and second-best Regret AUC values within each row are shown in bold and underlined, respectively.}
\label{tab:full_config_bo_results_101}
\scriptsize
\setlength{\tabcolsep}{3pt}
\renewcommand{\arraystretch}{1.15}
\resizebox{\textwidth}{!}{%
\begin{tabular}{@{}clcccccc@{}}
\toprule
\multicolumn{2}{c}{} & \multicolumn{4}{c}{Regret AUC ($\downarrow$)} & \multicolumn{2}{c}{TTARO vs. GP (\%)} \\
\cmidrule(lr){3-6}\cmidrule(lr){7-8}
Encoder & \shortstack{Kernel /\\Acquisition} & GP & GP-oracle & DKL & TTARO & \shortstack{Regret AUC\\Reduction} & \shortstack{FoM\\Improvement} \\
\midrule
\multirow{5}{*}{\rotatebox[origin=c]{90}{CktGNN}} & Linear / EI & $\underline{29770.0 \pm 355.2}$ & $30998.6 \pm 2389.1$ & $30955.8 \pm 707.8$ & $\boldsymbol{24216.1 \pm 1335.8}$ & $+18.7\%$ & $+11.8\%$ \\
 & Linear / TS & $27133.8 \pm 1029.9$ & $\boldsymbol{24201.1 \pm 725.0}$ & $29747.8 \pm 730.4$ & $\underline{24233.4 \pm 1308.2}$ & $+10.7\%$ & $+4.5\%$ \\
 & Linear / UCB & $\underline{27472.7 \pm 781.3}$ & $28508.1 \pm 2678.4$ & $30563.4 \pm 1154.1$ & $\boldsymbol{24161.9 \pm 1433.6}$ & $+12.1\%$ & $+12.4\%$ \\
 & RBF / EI & $27189.0 \pm 1297.9$ & $\boldsymbol{25945.1 \pm 1007.1}$ & $30056.9 \pm 500.5$ & $\underline{26311.4 \pm 1115.7}$ & $+3.2\%$ & $+3.3\%$ \\
 & RBF / UCB & $\boldsymbol{24013.6 \pm 1432.6}$ & $30246.8 \pm 2835.0$ & $28967.5 \pm 804.6$ & $\underline{25333.4 \pm 1107.9}$ & $-5.5\%$ & $-1.4\%$ \\
\addlinespace[2pt]
\hdashline
\addlinespace[2pt]
\multirow{5}{*}{\rotatebox[origin=c]{90}{WL}} & Linear / EI & $28160.2 \pm 206.1$ & $\underline{26814.7 \pm 591.0}$ & $30724.6 \pm 706.3$ & $\boldsymbol{15412.0 \pm 4556.1}$ & $+45.3\%$ & $+26.6\%$ \\
 & Linear / TS & $29498.1 \pm 361.6$ & $\underline{25101.0 \pm 1305.7}$ & $29766.7 \pm 492.2$ & $\boldsymbol{24445.5 \pm 2785.2}$ & $+17.1\%$ & $+14.1\%$ \\
 & Linear / UCB & $28502.1 \pm 242.4$ & $\underline{24795.8 \pm 1069.7}$ & $29394.9 \pm 556.0$ & $\boldsymbol{23400.4 \pm 5398.6}$ & $+17.9\%$ & $+12.8\%$ \\
 & RBF / EI & $32120.2 \pm 281.4$ & $\underline{27454.2 \pm 1000.2}$ & $29381.5 \pm 573.3$ & $\boldsymbol{21865.6 \pm 4634.5}$ & $+31.9\%$ & $+14.8\%$ \\
 & RBF / UCB & $\underline{30137.7 \pm 690.8}$ & $31957.3 \pm 2556.4$ & $30398.8 \pm 565.5$ & $\boldsymbol{22510.5 \pm 4905.0}$ & $+25.3\%$ & $+13.5\%$ \\
\addlinespace[2pt]
\hdashline
\addlinespace[2pt]
\multirow{5}{*}{\rotatebox[origin=c]{90}{D-VAE}} & Linear / EI & $51476.6 \pm 247.2$ & $\boldsymbol{18349.3 \pm 1671.7}$ & $36975.5 \pm 1670.0$ & $\underline{27449.0 \pm 1892.5}$ & $+46.7\%$ & $+30.8\%$ \\
 & Linear / TS & $41525.1 \pm 1395.0$ & $\boldsymbol{13801.4 \pm 1067.8}$ & $34733.7 \pm 1232.3$ & $\underline{28702.0 \pm 1548.7}$ & $+30.9\%$ & $+6.9\%$ \\
 & Linear / UCB & $49676.5 \pm 241.2$ & $\boldsymbol{14296.1 \pm 1102.2}$ & $33257.2 \pm 1061.2$ & $\underline{29071.5 \pm 1206.4}$ & $+41.5\%$ & $+10.4\%$ \\
 & RBF / EI & $\underline{29583.8 \pm 432.8}$ & $\boldsymbol{21193.7 \pm 1286.4}$ & $31867.2 \pm 1084.0$ & $30137.0 \pm 1221.2$ & $-1.9\%$ & $+5.7\%$ \\
 & RBF / UCB & $29862.4 \pm 946.9$ & $\boldsymbol{17272.9 \pm 2200.5}$ & $\underline{27824.2 \pm 1678.8}$ & $29766.0 \pm 1778.6$ & $+0.3\%$ & $+3.1\%$ \\
\addlinespace[2pt]
\hdashline
\addlinespace[2pt]
\multirow{5}{*}{\rotatebox[origin=c]{90}{D-VAE-GCN}} & Linear / EI & $39384.0 \pm 934.0$ & $\underline{29984.9 \pm 895.6}$ & $31186.6 \pm 486.6$ & $\boldsymbol{29766.5 \pm 436.9}$ & $+24.4\%$ & $+2.0\%$ \\
 & Linear / TS & $31408.0 \pm 622.8$ & $\boldsymbol{28338.1 \pm 316.7}$ & $31120.8 \pm 522.0$ & $\underline{29726.3 \pm 930.2}$ & $+5.4\%$ & $+2.2\%$ \\
 & Linear / UCB & $36606.0 \pm 940.1$ & $32457.1 \pm 2473.3$ & $\underline{30514.9 \pm 475.6}$ & $\boldsymbol{29297.6 \pm 402.3}$ & $+20.0\%$ & $+0.3\%$ \\
 & RBF / EI & $31107.1 \pm 415.8$ & $\boldsymbol{28540.1 \pm 836.0}$ & $31561.5 \pm 565.2$ & $\underline{30976.7 \pm 473.9}$ & $+0.4\%$ & $-0.3\%$ \\
 & RBF / UCB & $31925.0 \pm 624.3$ & $37901.6 \pm 3950.3$ & $\boldsymbol{30784.3 \pm 976.1}$ & $\underline{30928.8 \pm 555.2}$ & $+3.1\%$ & $+0.4\%$ \\
\addlinespace[2pt]
\midrule
\textit{Dataset Average} & \textit{All} & $32827.6 \pm 1645.5$ & $\boldsymbol{25907.9 \pm 1402.7}$ & $30989.2 \pm 459.3$ & $\underline{26385.6 \pm 882.5}$ & $+19.6\%$ & $+8.4\%$ \\
\bottomrule
\end{tabular}%
}
\end{table*}

\begin{table*}[ht]
\centering
\caption{Full configuration-level Bayesian optimization performance on Ckt-Bench-301. Each row corresponds to one encoder, kernel, and acquisition-function setting. Regret AUC is reported as mean $\pm$ standard error; lower is better. The best and second-best Regret AUC values within each row are shown in bold and underlined, respectively.}
\label{tab:full_config_bo_results_301}
\scriptsize
\setlength{\tabcolsep}{3pt}
\renewcommand{\arraystretch}{1.15}
\resizebox{\textwidth}{!}{%
\begin{tabular}{@{}clcccccc@{}}
\toprule
\multicolumn{2}{c}{} & \multicolumn{4}{c}{Regret AUC ($\downarrow$)} & \multicolumn{2}{c}{TTARO vs. GP (\%)} \\
\cmidrule(lr){3-6}\cmidrule(lr){7-8}
Encoder & \shortstack{Kernel /\\Acquisition} & GP & GP-oracle & DKL & TTARO & \shortstack{Regret AUC\\Reduction} & \shortstack{FoM\\Improvement} \\ 
\midrule
\multirow{5}{*}{\rotatebox[origin=c]{90}{CktGNN}} & Linear / EI & $22924.1 \pm 375.6$ & $\boldsymbol{1766.6 \pm 31.7}$ & $26903.2 \pm 593.8$ & $\underline{15114.6 \pm 1118.4}$ & $+34.1\%$ & $+27.0\%$ \\
 & Linear / TS & $18035.6 \pm 977.6$ & $\boldsymbol{1848.9 \pm 56.5}$ & $25663.6 \pm 444.9$ & $\underline{14526.3 \pm 1042.7}$ & $+19.5\%$ & $+3.9\%$ \\
 & Linear / UCB & $20303.5 \pm 847.0$ & $\boldsymbol{1766.6 \pm 31.7}$ & $25786.7 \pm 638.9$ & $\underline{14432.1 \pm 1232.0}$ & $+28.9\%$ & $+17.0\%$ \\
 & RBF / EI & $17409.3 \pm 1249.0$ & $\boldsymbol{3469.8 \pm 478.4}$ & $23175.6 \pm 599.4$ & $\underline{14037.8 \pm 1176.5}$ & $+19.4\%$ & $+5.9\%$ \\
 & RBF / UCB & $16808.7 \pm 1080.1$ & $\underline{14141.3 \pm 2534.5}$ & $22509.6 \pm 627.1$ & $\boldsymbol{11851.8 \pm 1078.7}$ & $+29.5\%$ & $+3.9\%$ \\
\addlinespace[2pt]
\hdashline
\addlinespace[2pt]
\multirow{5}{*}{\rotatebox[origin=c]{90}{DAGNN}} & Linear / EI & $12144.2 \pm 1253.1$ & $\boldsymbol{1977.0 \pm 153.5}$ & $18767.5 \pm 1639.5$ & $\underline{11183.7 \pm 1472.5}$ & $+7.9\%$ & $-0.0\%$ \\
 & Linear / TS & $12759.9 \pm 1509.4$ & $\boldsymbol{1860.2 \pm 52.6}$ & $21115.5 \pm 1042.7$ & $\underline{12016.2 \pm 1487.8}$ & $+5.8\%$ & $+2.0\%$ \\
 & Linear / UCB & $12397.7 \pm 1418.3$ & $\boldsymbol{2003.9 \pm 161.2}$ & $19685.3 \pm 1289.4$ & $\underline{9978.7 \pm 1195.3}$ & $+19.5\%$ & $+1.8\%$ \\
 & RBF / EI & $15874.3 \pm 1270.5$ & $\boldsymbol{3502.8 \pm 528.8}$ & $19441.8 \pm 1418.0$ & $\underline{11742.7 \pm 1315.5}$ & $+26.0\%$ & $+3.9\%$ \\
 & RBF / UCB & $17158.5 \pm 1237.5$ & $\underline{11857.9 \pm 2158.6}$ & $20436.6 \pm 1172.2$ & $\boldsymbol{11435.3 \pm 1413.4}$ & $+33.4\%$ & $+8.0\%$ \\
\addlinespace[2pt]
\hdashline
\addlinespace[2pt]
\multirow{5}{*}{\rotatebox[origin=c]{90}{D-VAE}} & Linear / EI & $23815.5 \pm 120.5$ & $\boldsymbol{22097.7 \pm 59.6}$ & $27820.9 \pm 1051.0$ & $\underline{23215.3 \pm 473.9}$ & $+2.5\%$ & $+5.6\%$ \\
 & Linear / TS & $23537.2 \pm 217.9$ & $\boldsymbol{19463.5 \pm 822.6}$ & $25414.4 \pm 497.4$ & $\underline{22248.8 \pm 775.2}$ & $+5.5\%$ & $+11.1\%$ \\
 & Linear / UCB & $24378.6 \pm 202.2$ & $\boldsymbol{17761.2 \pm 1078.4}$ & $26079.9 \pm 682.7$ & $\underline{23104.9 \pm 315.3}$ & $+5.2\%$ & $+8.5\%$ \\
 & RBF / EI & $23257.0 \pm 408.0$ & $\boldsymbol{21196.8 \pm 940.5}$ & $23375.2 \pm 606.5$ & $\underline{22638.4 \pm 852.9}$ & $+2.7\%$ & $+2.4\%$ \\
 & RBF / UCB & $22114.3 \pm 1043.6$ & $\underline{21640.2 \pm 1222.5}$ & $23506.1 \pm 651.2$ & $\boldsymbol{21583.1 \pm 898.6}$ & $+2.4\%$ & $-2.5\%$ \\
\addlinespace[2pt]
\hdashline
\addlinespace[2pt]
\multirow{5}{*}{\rotatebox[origin=c]{90}{D-VAE-GCN}} & Linear / EI & $\underline{23006.3 \pm 55.0}$ & $\boldsymbol{22590.7 \pm 75.4}$ & $24399.2 \pm 644.1$ & $23212.5 \pm 146.3$ & $-0.9\%$ & $+0.1\%$ \\
 & Linear / TS & $23131.8 \pm 81.1$ & $\boldsymbol{17468.6 \pm 683.2}$ & $25754.1 \pm 644.3$ & $\underline{21442.6 \pm 941.6}$ & $+7.3\%$ & $+14.0\%$ \\
 & Linear / UCB & $23117.0 \pm 195.8$ & $\boldsymbol{20126.8 \pm 1954.5}$ & $28162.3 \pm 1291.0$ & $\underline{21895.1 \pm 791.1}$ & $+5.3\%$ & $+13.9\%$ \\
 & RBF / EI & $23002.8 \pm 102.2$ & $23091.9 \pm 151.5$ & $\boldsymbol{22194.6 \pm 609.7}$ & $\underline{22348.7 \pm 575.2}$ & $+2.8\%$ & $+2.8\%$ \\
 & RBF / UCB & $22846.5 \pm 84.4$ & $24660.5 \pm 866.7$ & $\underline{22487.6 \pm 914.4}$ & $\boldsymbol{21605.2 \pm 552.8}$ & $+5.4\%$ & $+22.3\%$ \\
\addlinespace[2pt]
\midrule
\textit{Dataset Average} & \textit{All} & $19901.1 \pm 936.4$ & $\boldsymbol{12714.6 \pm 2061.5}$ & $23634.0 \pm 628.4$ & $\underline{17480.7 \pm 1149.2}$ & $+12.2\%$ & $+7.2\%$ \\
\bottomrule
\end{tabular}%
}
\end{table*}

\subsection{Optimization Protocol and Evaluation Metrics}
All methods are evaluated under the same finite-bank BO protocol. For a given benchmark, encoder, acquisition function, and random seed, the methods begin from the same initial evaluated circuit set. At each BO iteration, the surrogate is fit using the FoM labels observed so far, the acquisition function is evaluated over the remaining unevaluated candidates, and the selected candidate's stored FoM is revealed and added to the observed set. Except for the oracle representation condition, no method uses FoM labels from unevaluated candidates when fitting its representation or surrogate. We evaluate two GP kernels, a linear kernel and an RBF kernel, together with five kernel/acquisition settings: Linear/EI, Linear/TS, Linear/UCB, RBF/EI, and RBF/UCB.

We evaluate performance using best-so-far FoM and regret-based metrics, which are standard in BO and GP bandit optimization \cite{srinivas2009gaussian_UCB, shahriari2015taking}. Let
\begin{equation}
    f^\star =
    \max_{i \in \{1,\ldots,N\}} y_i
\end{equation}
denote the best FoM in the candidate bank, used only for evaluation. After $t$ circuit evaluations, the best observed FoM is
\begin{equation}
    f_t^+
    =
    \max_{i \in \mathcal{I}_t} y_i .
\end{equation}
The simple regret at iteration $t$ is
\begin{equation}
    r_t
    =
    f^\star - f_t^+ .
\end{equation}
Lower regret indicates that a method has found higher-performing circuits earlier in the search.

We also report Top-2\% KNN@5 in the representation-geometry visualization (see Fig. \ref{fig:ttaro_representation_dynamics}) to help interpret the UMAP plots. For each circuit whose FoM is in the top 2\% of the candidate bank, Top-2\% KNN@5 measures the fraction of its five nearest representation-space neighbors that are also top-2\% FoM circuits. Higher values indicate that high-performing circuits are more locally concentrated in the learned representation. This quantity is used only as a diagnostic for the visualization; the primary optimization metrics are regret AUC and final best-so-far FoM.

To summarize the full optimization trajectory, we report regret AUC over the evaluation budget. For a budget of $B$ evaluations, regret AUC is computed as the discrete area under the simple-regret curve:
\begin{equation}
    \operatorname{AUC}_{r}
    =
    \sum_{t=1}^{B-1}
    \frac{r_t + r_{t+1}}{2}.
\end{equation}
Equivalently, this is the trapezoidal approximation to the regret curve when evaluations are spaced one step apart. Lower regret AUC indicates better sample efficiency across the full BO run. We also report final best-so-far FoM to measure the best circuit found by the end of the budget.

For TTARO, regret AUC reduction over GP is reported as
\begin{equation}
    \Delta_{\mathrm{AUC}}
    =
    \frac{
        \mathrm{AUC}_{r}^{\mathrm{GP}}
        -
        \mathrm{AUC}_{r}^{\mathrm{TTARO}}
    }{
        \mathrm{AUC}_{r}^{\mathrm{GP}}
    }
    \times 100 .
\end{equation}
Final FoM improvement over GP is reported as
\begin{equation}
    \Delta_{\mathrm{FoM}}
    =
    \frac{
        \mathrm{FoM}_{B}^{\mathrm{TTARO}}
        -
        \mathrm{FoM}_{B}^{\mathrm{GP}}
    }{
        \mathrm{FoM}_{B}^{\mathrm{GP}}
    }
    \times 100 .
\end{equation}

\section{Results and Discussion} \label{sec:conclusion}


Tables~\ref{tab:full_config_bo_results_101} and~\ref{tab:full_config_bo_results_301} summarize the configuration-level regret AUC results. TTARO improves regret AUC relative to fixed-representation GP on both Ckt-Bench-101 and Ckt-Bench-301, indicating that online representation adaptation improves the sample efficiency of BO. On Ckt-Bench-101, TTARO reduces the dataset-average regret AUC from $32827.6$ to $26385.6$, a $19.6\%$ reduction relative to GP. On Ckt-Bench-301, TTARO reduces the dataset-average regret AUC from $19901.1$ to $17480.7$, a $12.2\%$ reduction. These gains are consistent with the intended role of TTARO: as FoM observations accumulate, the surrogate can update the kernel geometry used to generalize from evaluated to unevaluated circuits.

The comparison with DKL highlights why online adaptation matters. On Ckt-Bench-101, DKL improves regret AUC relative to GP, while its final best FoM remains slightly lower than GP. On Ckt-Bench-301, DKL degrades substantially: regret AUC increases from $19901.1$ for GP to $23634.0$, and final best FoM drops from $152.7$ to $134.0$. This suggests that a representation learned only from the initial evaluated set can be brittle, especially in the larger and more heterogeneous search space. TTARO avoids this degradation, reducing regret AUC by $14.9\%$ relative to DKL on Ckt-Bench-101 and by $26.0\%$ on Ckt-Bench-301.

The configuration-level results in Tables~\ref{tab:full_config_bo_results_101} and~\ref{tab:full_config_bo_results_301} show that these aggregate gains are broadly distributed across encoders, kernels, and acquisition functions. TTARO improves regret AUC relative to GP in 37 of the 40 evaluated settings. The three exceptions are CktGNN/RBF/UCB and D-VAE/RBF/EI on Ckt-Bench-101, where TTARO increases regret AUC relative to GP by $5.5\%$ and $1.9\%$, respectively, and D-VAE-GCN/Linear/EI on Ckt-Bench-301, where TTARO increases regret AUC by $0.9\%$.

Table~\ref{tab:fom_by_budget} reports aggregate best-so-far FoM at fixed fractions of the BO budget. This view complements regret AUC by showing how quickly each method finds high-performing circuits during the search. On Ckt-Bench-101, TTARO reaches $228.6$ FoM by $20\%$ of the budget, compared with $210.4$ for GP and $221.2$ for DKL. By $60\%$ of the budget, TTARO reaches $252.8$, already exceeding the final GP value of $247.6$. On Ckt-Bench-301, TTARO reaches $152.8$ by $60\%$ of the budget, matching the final GP value of $152.7$ while still having $40\%$ of the evaluation budget remaining. These budgeted FoM trends show that TTARO's regret-AUC gains correspond to earlier discovery of high-performing circuits, not only lower integrated regret.

The oracle representation condition provides useful context for interpreting the upper end of representation quality. It trains the feature map using all FoM labels in the candidate bank before BO begins, giving the surrogate access to the global objective landscape. This allows the learned representation to place circuits with similar FoM values closer together and separate circuits with substantially different FoM values before any sequential search decisions are made. As a result, the GP kernel begins with a more objective-aligned covariance structure, so observations from evaluated circuits can be transferred more effectively to unevaluated candidates. \textbf{Since the oracle representation is trained using FoM labels that are unavailable during a real BO run, it is not a deployable optimization method and should be interpreted only as an upper-reference condition.}

Figure~\ref{fig:auc_encoder} summarizes these configuration-level trends by kernel and acquisition setting. On Ckt-Bench-301, TTARO improves over GP in all but one encoder/kernel/acquisition setting, and every kernel/acquisition family improves on average across encoders. The gains are especially important for the linear-kernel settings, where improving the representation directly changes the inner-product geometry used by the surrogate. TTARO also improves the RBF settings, showing that the method is useful when the kernel uses a nonlinear distance-based form. The detailed results also show different trends across kernels. On Ckt-Bench-101, the largest TTARO gains occur in the linear-kernel settings, with average regret AUC reductions of $33.8\%$ for Linear/EI, $16.0\%$ for Linear/TS, and $22.9\%$ for Linear/UCB. The RBF settings are more mixed on Ckt-Bench-101, although TTARO still improves most of them. On Ckt-Bench-301, the average gains are $10.9\%$ for Linear/EI, $9.5\%$ for Linear/TS, $14.7\%$ for Linear/UCB, $12.7\%$ for RBF/EI, and $17.7\%$ for RBF/UCB.

\begin{table*}[hbt]
\centering
\caption{Aggregate best-observed FoM across Ckt-Bench-101 and Ckt-Bench-301 over the BO evaluation budget. Mean $\pm$ standard error is reported at 20\% increments of the budget; higher is better. The best value within each dataset and budget checkpoint is shown in bold, and the second-best value is underlined.}
\label{tab:fom_by_budget}
\tiny
\setlength{\tabcolsep}{3pt}
\renewcommand{\arraystretch}{1.15}
\resizebox{\textwidth}{!}{%
\begin{tabular}{@{}clcccc>{\columncolor{gray!15}}c@{}}
\toprule
Dataset & Method &
\shortstack{Best FoM\\@ 20\%} &
\shortstack{Best FoM\\@ 40\%} &
\shortstack{Best FoM\\@ 60\%} &
\shortstack{Best FoM\\@ 80\%} &
\shortstack{Best FoM\\@ 100\%} \\
\midrule

\multirow{4}{*}{\rotatebox[origin=c]{90}{\shortstack{Ckt-Bench\\101}}} & GP & $210.4 \pm 2.0$ & $225.9 \pm 1.5$ & $231.8 \pm 1.6$ & $237.6 \pm 1.5$ & $247.6 \pm 1.3$ \\
 & GP-oracle & $\boldsymbol{231.2 \pm 2.0}$ & $\boldsymbol{249.0 \pm 2.1}$ & $\boldsymbol{255.0 \pm 2.1}$ & $\boldsymbol{261.9 \pm 2.3}$ & $\boldsymbol{271.0 \pm 2.3}$ \\
 & DKL & $221.2 \pm 1.4$ & $234.1 \pm 0.8$ & $238.5 \pm 0.8$ & $241.7 \pm 0.8$ & $243.5 \pm 0.8$ \\
 & TTARO & $\underline{228.6 \pm 1.6}$ & $\underline{245.3 \pm 1.4}$ & $\underline{252.8 \pm 1.6}$ & $\underline{258.2 \pm 1.8}$ & $\underline{264.6 \pm 2.0}$ \\

\midrule
\multirow{4}{*}{\rotatebox[origin=c]{90}{\shortstack{Ckt-Bench\\301}}} & GP & $124.7 \pm 0.7$ & $132.3 \pm 1.0$ & $141.7 \pm 1.5$ & $147.2 \pm 1.6$ & $152.7 \pm 1.7$ \\
 & GP-oracle & $\boldsymbol{153.6 \pm 2.0}$ & $\boldsymbol{157.8 \pm 1.9}$ & $\boldsymbol{161.6 \pm 1.9}$ & $\boldsymbol{167.9 \pm 1.8}$ & $\boldsymbol{171.8 \pm 1.8}$ \\
 & DKL & $110.7 \pm 1.1$ & $122.2 \pm 0.9$ & $128.5 \pm 1.0$ & $131.9 \pm 1.1$ & $134.0 \pm 1.2$ \\
 & TTARO & $\underline{127.0 \pm 0.9}$ & $\underline{141.9 \pm 1.5}$ & $\underline{152.8 \pm 1.7}$ & $\underline{158.5 \pm 1.8}$ & $\underline{163.7 \pm 1.8}$ \\

\bottomrule
\end{tabular}%
}
\end{table*}

The main limitation of TTARO is computational overhead. Retraining the feature map and GP after each evaluation is more expensive than updating a fixed-representation GP posterior, and this cost may matter for very large candidate banks or very short simulation times. Existing scalable GP and deep-kernel methods provide several paths for reducing this overhead. Structured kernel interpolation reduces the cost of kernel-matrix operations and enables scalable kernel learning \cite{kissGP}; stochastic variational deep-kernel learning supports scalable joint training of neural feature maps and GP kernels \cite{stochasticVariationalKernelLearning}; Lanczos-based variance estimation and GPU-accelerated GP inference reduce the cost of posterior uncertainty computation and sampling \cite{constantTimePredictiveGP,gpytorch}. Online GP methods further address the sequential-update setting by reusing computations after new observations arrive \cite{kernelInterpolation}. These methods suggest that future TTARO implementations can reduce surrogate-update cost through structured kernels, sparse variational approximations, and GPU-accelerated linear algebra. In the analog design regimes targeted here, circuit evaluation is typically the dominant cost, making additional surrogate training acceptable when it reduces the number of poor evaluations.

Overall, the results support the central premise of this work: for representation-based circuit BO, the quality of the surrogate depends strongly on whether the kernel geometry is aligned with the objective being optimized. Fixed GP relies entirely on the initial circuit embedding, and DKL learns a transformation only once from the initial data. TTARO keeps the representation coupled to the online search process. This lets the surrogate revise its notion of circuit similarity as new FoM evidence becomes available, which leads to better acquisition decisions and improved sample efficiency.

\section*{Acknowledgments}
The authors used ChatGPT-5.5 for grammatical refinements, boilerplate code generation and to improve the aesthetic aspects of the figures. We assume full responsibility for all the content in this manuscript. We would like to thank the authors of OCB for making their code and datasets available. 

\balance
\bibliographystyle{IEEEtran}
\bibliography{main}

\end{document}